\documentclass{article}

\usepackage{arxiv}

\usepackage[utf8]{inputenc} 
\usepackage[T1]{fontenc}    
\usepackage{hyperref}       
\usepackage{url}            
\usepackage{booktabs}       
\usepackage{amsfonts}       
\usepackage{nicefrac}       
\usepackage{microtype}      
\usepackage{lipsum}		
\usepackage{graphicx}
\usepackage[square, comma, sort&compress, numbers]{natbib}
\usepackage{doi}
\usepackage{amsmath}
\usepackage{algorithm}
\usepackage{algorithmic}
\usepackage{cite}

\title{Extreme-Long-Short Term Memory for Time-series Prediction}

\date{} 					

\author{ Sida Xing\\
    Deakin University\\
    \texttt{sxing@deakin.edu.au} \\
    \And
    Feihu Han\\
    Deakin University\\
    \texttt{feihuh@deakin.edu.au}
    \And
    Dr Sui Yang Khoo\\
    Deakin University\\
    \texttt{sui.khoo@deakin.edu.au}
}

\renewcommand{\headeright}{}
\renewcommand{\undertitle}{}
\renewcommand{\shorttitle}{\textit{E-LSTM}}

\hypersetup{
pdftitle={Extreme-Long-Short Term Memory for Time-series Prediction},
pdfauthor={Sida Xing, Feihu Han, Sui Yang Khoo},
}

\begin{document}
\maketitle

\begin{abstract}
The emergence of Long Short-Term Memory (LSTM) solves the problems of vanishing gradient and exploding gradient in traditional Recurrent Neural Networks (RNN). LSTM, as a new type of RNN, has been widely used in various fields, such as text prediction \citep{buddana2021word}, Wind Speed Forecast \citep{moharm2020wind} and depression prediction by EEG signals, etc. \citep{kumar2019prediction}. The results show that improving the efficiency of LSTM can help to improve the efficiency in other application areas \citep{christiaanse1971short}.

In this paper, we proposed an advanced LSTM algorithm, the Extreme Long Short-Term Memory (E-LSTM), which adds the inverse matrix part of Extreme Learning Machine (ELM) as a new "gate" into the structure of LSTM. This "gate" preprocess a portion of the data and involves the processed data in the cell update of the LSTM to obtain more accurate data with fewer training rounds, thus reducing the overall training time.

In this research, the E-LSTM model is used for the text prediction task. Experimental results showed that the E-LSTM sometimes takes longer to perform a single training round, but when tested on a small data set, the new E-LSTM requires only 2 epochs to obtain the results of the $7^{th}$ epoch of the LSTM. Therefore, the E-LSTM retains the high accuracy of the traditional LSTM, whilst also improving the training speed and the overall efficiency of the LSTM.
\end{abstract}

\section{Introduction}
In 1988, Boggess and Lois worked on the input speed problem for people with disabilities by using two simple text prediction algorithms to improve input speed \citep{boggess1988two}. Trnka et al. (2006) agreed that approximately 2 million people in the United States had some form of communication difficulty in 2006. ACC's device developers were unable to improve user input rates, and therefore needed to improve input speed by reducing the amount of input for common tasks to improve communication output \citep{trnka2006topic}. Ayodele and Zhou (2008) proposed that unsupervised learning can be used to analyze emails and help people better organize and prioritize emails by analyzing the content features \citep{ayodele2008email}. 

Traditional neural networks were unable to find out contextual associations of predictions when dealing with predicted sentence words \citep{lippmann1994book}. RNN has the memory of previous time states, which helps the system acquire context. Theoretically, RNNs have time series memory, which means it has good performance for time series problems. However, in practical terms, it can only review the last few steps \citep{lippmann1994book, sherstinsky2020fundamentals}.
 
With the further use of recurrent neural networks, Hochreiter, Sepp \& Schmidhuber, Jürgen. (1997) agreed that recurrent neural networks are widely used in short-term memory processing, but not ideal for long-term processing \citep{hochreiter1997long}.

During the backpropagation of recurrent neural networks, RNN have the problem of gradient vanishing and gradient exploding when dealing with long-term sequence learning, which will directly affect the learning performance of the model \citep{hochreiter1997long}.

Therefore, LSTM was born to solve the vanishing gradient and exploding gradient problems for the recurrent neural network. Although LSTM was created many years ago, there are still many people using it in different today. Such as text prediction \citep{buddana2021word, santhanam2020context}, Wind Speed Forecast\citep{moharm2020wind, geng2020short} and depression prediction by EEG signals, etc.\citep{kumar2019prediction}.

In the training of the traditional neural network, the usual practice is to continuously adjust the hidden layer and the output layer, the weight matrix between the output layer and the hidden layer, and the bias ‘b’ through the gradient descent algorithm. This means that all parameters in traditional feedforward neural network need to be adjusted, making the neural networks take more time to adjust and the parameter correlated in different layers\citep{huang2004extreme, yu2019review}.

However, the ELM pointed out that it is unnecessary to adjust the weight matrix ‘W’ and the bias ‘b’ of the hidden layer. In the beginning, the values of "W" and "b" are given randomly, and "H" (hidden layer node output) is calculated and held constant by using random weights and deviations, and the only parameter that needs to be determined is $\beta$\citep{huang2004extreme}.

In this paper, we used the Extreme Learning Machine (ELM) to optimize the LSTM, with fast learning characteristics. By adding a new "gate" to the LSTM, the fast-training results of the ELM can participate in the update process of the LSTM cells, thus The LSTM can reduce the number of iterations of the LSTM by combining the results obtained from the ELM in the memory process and filtering the information by the activation function. Based on the combined framework of ELM and LSTM, we designed a novel E-LSTM model. The new model will be using for text prediction.

\section{Related Work}

\subsection{Long-short term memory (LSTM)}
The structure of the recurrent neural network is composed of three layers, there are input layer, hidden layer and output layer. In the process of forward propagation, the current layer calculates the input data according to the network connections and the weights, then outputs to the next layer. Therefore, the recurrent neural network has an additional memory function compared to the BP neural network, giving it a greater advantage in time series information \citep{hochreiter1997long,yu2019review}.

When data information is stored for a long time, it may lead to gradient vanishing or gradient explosion. The reason is that when the information is transmitted at the input layer, the information is transmitted back in time will form an effective memory\citep{hochreiter1997long, chen2015lstm}. However, propagating errors that occur at unit 'u' with time step 't' back to unit 'v' with time step 'q' scales the errors"\citep{hochreiter1997long, yu2019review}.

The LSTM has been upgraded based on the original RNN. The upgraded RNN, which is LSTM network consists of input gate, output gate, forget gate and CELL. CELL is the core of the compute node and is used to record the state of the current time step \citep{yu2019review, gers2000learning}.

To establish the temporal connections, the LSTM defines and maintains the internal memory cell state-cell state $C_t$ throughout the cycle. Then it passes through the forget gate $f_t$ the input gate $i_t$, and the output gate $O_t$  to update information within maintenance or t cell status \citep{hochreiter1997long, karim2017lstm}.

\begin{figure}[ht]
	\centering
	\includegraphics[width=5in]{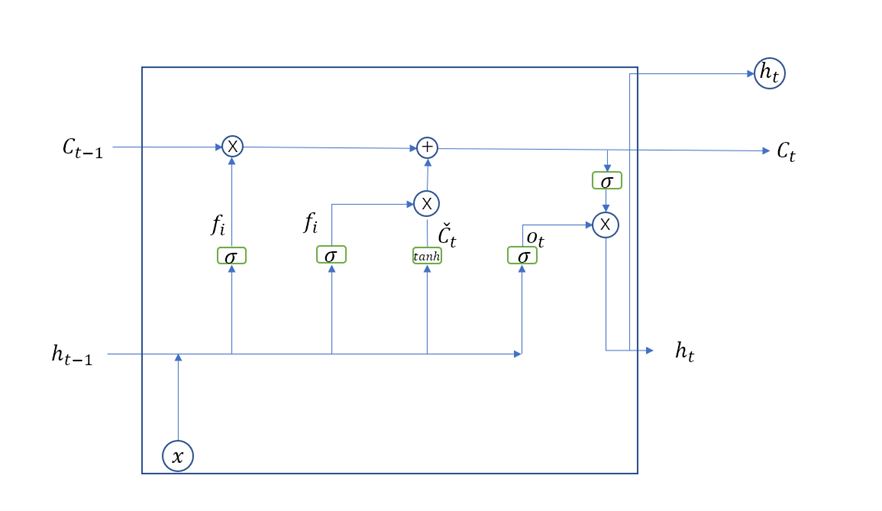}
	\caption{LSTM neural networks}
	\label{fig_LSTM}
\end{figure}

The process of forwarding calculation in LSTM is as follows[9]:

t represents the current time, and t-1 represents the previous time. 

$C_t$ represents the state of the Cell.

W represents a matrix of weight coefficients.

b represents the bias term.

$\sigma$ represents the sigmoid activation function.

tanh represents the hyperbolic tangent activation function [9].

\begin{equation}
    f_t=\sigma(W_f*[h_{t-1},x_t]+b_f),
\end{equation}

\begin{equation}
    i_t=\sigma(W_i*[h_{t-1},x_t]+b_i),
\end{equation}

\begin{equation}
    \widetilde{C_t}= tanh(W_c*[h_{t-1},x_t]+b_C),
\end{equation}

\begin{equation}
    C_t=f_t*C_{t-1}+i_t*(C_t),
\end{equation}

\begin{equation}
    o_t=\sigma(W_o*[h_{t-1},x_t]+b_o),
\end{equation}

\begin{equation}
    h_t=o_t*tanh(C_t)
\end{equation}

The hidden output $h_{(t-1)}$ and the current output $h_t$ at the last moment are calculated by equations 1, 2, and 5 to derive the forget gate \citep{hochreiter1997long, yu2019review}.

The input gate and the output gate coefficient, then, the state $C_t$ of the current neural can be obtained by equation 3, and then the ratio of $C_{(t-1)}$ and $C_t$ in the current cell, state is determined from the forget gate and the input gate, then completed by equation 4 \citep{hochreiter1997long}.

The state of the Cell is now updated. Finally, the output value of the hidden layer at the current time is calculated by equation 6 \citep{hochreiter1997long}.

The LSTM is trained by time series backpropagation for this network. The gradient of weight is updated by calculating the error \citep{hochreiter1997long}.

Hochreiter (1981) showed that “propagating back an error occurring at a unit u at time step t to a unit v for q time steps, scales the error ” by \citep{hochreiter1997long}:

\begin{equation}
    \frac{\partial \vartheta_v}{\partial\vartheta_u(t)}= 
    \left \{ 
    \begin{array}{c}
    f'_v(net_v(t-1)W_{uv}), q=1 \\ 
    f'_v(net_v(t-q))\sum_{i=1}^{n}{\frac{\partial\vartheta_l(t-q+1)}{\partial\vartheta_u(t)}}W_{lv}, q>1 
    \end{array}
    \right.
\end{equation}

with $l_{(q)}$ = v and $l_{(0)}$ = u, the scaling factor is\citep{hochreiter1997long}:

\begin{equation}
    \frac{\partial \vartheta_v(t-q)}{\partial\vartheta_u(t)}=\sum_{l_1=1}^{n}...\sum_{l_{q-1}=1}^{n}\prod_{m=1}^{q}f'_{lm}(net_v(t-m))wl_ml_{M-1}
\end{equation}

This equation shows that the absolute value of the error return will be larger when there is a more significant number of units. When m is less than a certain value, it will exponentially decrease and eventually approach 0, which is how it causes the gradient to vanish \citep{hochreiter1997long}.

Its calculation formula is:

\begin{equation}
    a_c(t) = \sum_t^lx_t(t)w_{ic}+\sum_h^H{b_h(t-1)w_{hc}}
\end{equation}

Hochreiter, Sepp \& Schmidhuber, Jürgen. (1997) agreed that LSTM combined with appropriate gradient algorithms is used for long-term memory, and that LSTM can perform undelayed work in the presence of input sequences that interfere with other factors. This also shows that LSTM can work more objectively and accurately\citep{hochreiter1997long, fu2016using}.

\subsection{Extreme learning machine (ELM)}
The connection weights of the input layer and the hidden layer, and the threshold of the hidden layer can be set randomly, but no adjustment is required after setting. This is different from the BP neural network, where BP needs to adjust the weights and thresholds in reverse constantly. Therefore, the computational effort here can be reduced by half\citep{huang2004extreme,huang2015extreme}.

The connection weight $\beta$ between the hidden layer and the output layer does not need to be adjusted iteratively, but is determined once by solving the equations\citep{huang2004extreme,huang2015extreme}.

Following Huang’s theory, the structure of ELM shows below\citep{huang2004extreme,huang2015extreme}:

\begin{figure}[ht]
	\centering
	\includegraphics[width=4in]{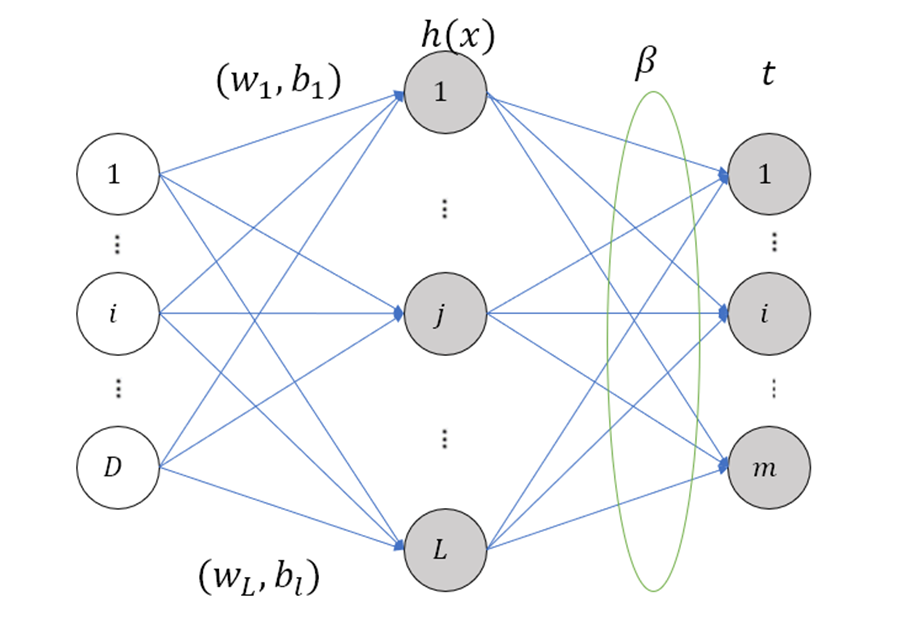}
	\caption{The structure of ELM}
	\label{fig_ELM}
\end{figure}

Figure \ref{fig_ELM} shows the input layer on the left. X is the input data. It is fully connected between the input layer and the hidden layer. The hidden layer node is L. H(X) is the output of the hidden layer and since the input X is the training sample set, the output H(X) is available\citep{huang2004extreme,huang2015extreme}:

\begin{equation}
    H(X)=[h_1(x),...h_L(X)]
\end{equation}

Figure \ref{fig_ELM} also illustrates that the output matrix of the hidden layer needs to be multiplied by the initial weight w and the bias value b. Thus, the output of a single neuron node in the hidden layer $h_i$(X), i is the neuron node\citep{huang2004extreme,huang2014optimization}.

\begin{equation}
    h_i(X)=g(w_i,b_i,X) = g(w_iX+b_i), w_i\in {R^D},{b_i\in{R}}
\end{equation}

The g is the activation function. Different activation functions can be used according to the situation, such as sigmoid, tanh and ReLU activation functions. Assuming g is the sigmoid activation function\citep{huang2004extreme}:

\begin{equation}
    g(X)=\frac{1}{1+e^{-x}}=\frac{e^x}{e^x+1}
\end{equation}

Next is the hidden layer to the output layer. According to the formula and image, the output of the ELM neural network is\citep{huang2004extreme}:

\begin{equation}
    f_L(X)=\sum^{L}_{i=1}\beta_i{h_i}(x)=H(X)\beta
\end{equation}

$\beta$ in the formula is the weight matrix from the hidden layer to the output layer, and $\beta_i$ is the output weight of a single neuron from the hidden layer to the output layer\citep{huang2004extreme,huang2015extreme}.

Although SLFN (single-layer feedforward neural network) is based on ELM, the performance of ELM is much better than SLFN because ELM has a special algorithm.\citep{huang2004extreme,wang2015state}.

Training SLFN in ELM is mainly divided into two stages: random feature mapping and linear parameter solving\citep{huang2004extreme}.

In the first stage, the hidden layer parameters w and b are randomly initialized, and then the input data are mapped to a new feature space using a nonlinear mapping as the activation function\citep{huang2004extreme}.

After initializing the parameters w and b in the first stage, the hidden layer output H can be calculated\citep{huang2004extreme}.

By the above method, with single hidden layer output $H\beta$ and sample label T, the minimum squared difference is obtained as the evaluation training error, and the best $\beta$ can be obtained by finding the solution with the minimum training error. i.e., the method of minimizing the approximate squared difference and solving for the weight $\beta$ connecting the hidden layer and the output layer\citep{huang2004extreme}:

\begin{equation}
    min \parallel H\beta-T\parallel^2,\beta \in R^{L\times M}
\end{equation}

where, H is the hidden layer output matrix, and T is the training target matrix\citep{huang2004extreme,huang2014optimization}:

\begin{equation}
    H=[h(x_1),...,h(x_N)]^T = 
    \begin{bmatrix}
        H_1(X_1)&\cdots&H_L(X_1) \\
        \vdots&\ddots&\vdots \\
        H_1(X_n)&\cdots&H_L(X_N) \\
    \end{bmatrix},
    T=\begin{bmatrix}
        t_1^t \\
        \vdots \\
        t_N^T \\
    \end{bmatrix}
\end{equation}

This can be obtained by formula inference\citep{huang2004extreme,huang2014optimization}:

\begin{equation}
    \beta^* = H^\dag T
\end{equation}

where $H^\dag$is the Moore-Penrose generalized inverse of matrix H\citep{huang2004extreme}.

At this time, the problem is transformed into calculating the Moore-Penrose generalized inverse matrix of the matrix H\citep{huang2004extreme}.

The main methods for this problem are orthogonal projection, orthogonalization, iteration, and singular value decomposition (SVD). When $H^T$H orthogonal projection method is available for non-singular (invertible), the available calculations are\citep{huang2004extreme}:

\begin{equation}
    H^\dag = (H^TH)^{-1}H^T
\end{equation}

After proposing the extreme learning machine algorithm, ELM was applied to events such as house price prediction, diabetes testing and forest cover type prediction. After that, other authors applied ELM for different purposes\citep{huang2004extreme}.

The research of Y.Cui et al. (2016) shows the advantages and disadvantages of LSTM and ELM\citep{cui2016comparative}. The advantage of LSTM is that it can be trained for long-short term with high accuracy. The accuracy of ELM is lower than LSTM, but the calculation speed is faster than LSTM.

In the previous section, it can be observed that LSTM is multilayer neural network and ELM is single layer feed-forward neural network. Therefore, the LSTM structure is more complicated than ELM in data processing\citep{hochreiter1997long,huang2004extreme}.

G.-B. Huang et al.(2004) shows that initial weights and bias values are randomized of ELM as follows the function\citep{huang2004extreme}:

\begin{equation}
    h_i(X)=g(w_i,b_i,X) = g(w_iX+b_i), w_i\in {R^D},{b_i\in{R}}]
\end{equation}

to get the output of the hidden layer which is the result of  H(x)\citep{huang2004extreme}.

In function 26, g is the activation function, it can be sigmoid, ReLU or other activation function. ELM through simple generalized inverse of matrix to get the output weight of the hidden layer, which as follows the formula\citep{huang2004extreme}: 

\begin{equation}
    \beta^* = H^\dag T
\end{equation}

The LSTM has three gates, there are forget gate, input gate and output gate. The function of the forget gate as follows\citep{hochreiter1997long,gers2000learning}:

\begin{equation}
    f_t = \sigma(W_f*[h_{t-1},x_t]+b_f)
\end{equation}

As can be seen from equations (19) and (21), when the activation function sigmoid is used for the hidden layer in ELM. Hidden layer output of ELM $h_i$ (X) equal to the output of forget gate in LSTM. Because of this, through Moore-Penrose generalized inverse of matrix can calculate the output weight of forget gate in LSTM, which is as follows the equation (20)\citep{hochreiter1997long,huang2004extreme}.

In this research, will adding the generalized inverse of matrix part in ELM to LSTM be a new gate called E-gate. The E-gate will process the input data as ELM, which means calculate the input data ($h_(t-1)$,$x_t$) to get weight with accuracy, then combine the weight with forget gate and input gate to make a cell. Finally, the cell performs iterations. This will make the data of the cell have higher accuracy than before, which means the E-LSTM can achieve the same accuracy as LSTM with fewer training rounds.

When a new gate adds to LSTM, a new structure LSTM was created called Extreme long-short-term memory(E-LSTM).

The process of E- LSTM as follows:

t represents the current time, and t-1 represents the previous time.

C represents the state of the Cell.

W represents a matrix of weight coefficients.

$f_t$ represents the forget gate.

$i_t$ represents the input gate.

$o_t$ represents the output gate.

b represents the bias.

$e_t$ represents the E-gate.

T represents the target matrix for the E-gate.

$h_t$ represents the output.

tanh represents the hyperbolic tangent activation function.

$\sigma$ represents the sigmoid activation function.

Assume that given a data set N = [($x_i$, $t_i$) | $x_i$ $\in$ $R^n$, t $\in$ $R^m$, i = 1, ... N]

Step1: Randomly assign input weight W and bias b  of the forget gate, input gate and output gate.

Step2: Calculate the output of the forget gate $f_t$ and input gate $i_t$ as follows: 

\begin{equation}
    f_t = \sigma(W_f*[h_{t-1,x_t}]+b_f)
\end{equation}

\begin{equation}
    i_t = \sigma(W_i*[h_{t-1,x_t}]+b_i)
\end{equation}

Step3: Calculate the output of the E-gate $e_t$ as follows:

\begin{equation}
    e_t = f^\dag T
\end{equation}

Where T=[$t_1$,…$t_N$]$^N$,\citep{huang2004extreme}

Where $f^\dag$ is the Moore-Penrose generalized inverse of matrix forget gate $f_t$.

Step4: Calculate the output of the Cell$\_$bar $\widetilde{C_t}$ as follows:

\begin{equation}
    \widetilde{C_t}= tanh(W_c*[h_{t-1},x_t]+b_c),
\end{equation}

Step5: Calculate the output of the Cell $C_t$  as follows:

\begin{equation}
    C_t=f_t*C_{t-1}+i_t*\widetilde{C}_t+e_t,
\end{equation}

Step6: Calculate the output of the output gate $o_t$  as follows:

\begin{equation}
    o_t=\sigma(W_o*[h_{t-1},x_t]+b_o)
\end{equation}

Step7: Calculate the output of the final networks $h_t$  as follows:

\begin{equation}
    h_t=o_t*tanh(C_t)
\end{equation}

The internal structure of the E-LSTM neural networks is shown in formulas (22) to (28). 
As shown in the process of the E-LSTM, the structure of E-LSTM is similar to LSTM because the E-LSTM needs the three gates of LSTM to make long-term memory calculations\citep{hochreiter1997long,huang2004extreme}. 

The input gate, forget gate and output gate in E-LSTM work the same as in LSTM. Forget gate map the output to the interval [0,1] by using the activate function sigmoid, the data map to the 0 will drop out, then other data multiply with the previous state cell, and then the total output enters the current state cell to participate in the next calculation. The role of the input gate is to decide whether the output of the hidden layer at the previous time state and the input at the current time state are retained through the sigmoid function. The output gate is to decide the output of the hidden layer in the previous time state and the input of the current time state\citep{hochreiter1997long,huang2004extreme}.

The difference between LSTM and E-LSTM is the output of the cell. The formula (26) shows the E-gate has been added to the cell $C_t$. The role of E-gate is to further calculate the output of the forget gate before multiplying it with the previous time state cell and calculating the weight by generalized inverse of matrix, which has higher accuracy than the Initial input data. In other words, E-gate is equivalent to a mini single layer neural network filtering the input data to make the data has contains accurate data then added to the cell iterations, and this forget gate retains more useful information when the data passes through the forget gate ft+1 of the next time state, thus reducing the number of training rounds\citep{hochreiter1997long,huang2004extreme}. 

Although E-gate is further calculated by the initial output weight and bias value of the forget gate, it does not mean that E-gate can totally replace the forget gate. The main function of the forget gate is to retain the useful information in the input data, while the main function of the E-gate is to filter the input data further and improve the accuracy of the input data.

This structure makes E-LSTM keep the long and short-term data processing features of LSTM and accuracy higher than LSTM in single epoch calculation.

In theory, the structure of E-LSTM would be more complex compared to LSTM because E-LSTM needs two more steps than LSTM in processing the information, firstly, it needs to get the output in E-gate by the generalized inverse of matrix and the target matrix, and then add the output to Cell at the current time state. And E-LSTM would require more time to process the data in a single training count. However, the total training time of E-LSTM will be less than that of traditional LSTM, and the efficiency of the neural network will be improved by decreasing the total training time.

\section{Experiments}

\subsection{Language Modeling}
The whole model was built using the python 3.7.7 language. To approach the LSTM theoretical model, the whole model will be built based on the numpy frame. The experimental equipment used comes with a 3090 graphics card and a 3970X thread stripper CPU.

In this research, E-LSTM and traditional LSTM are used to make predictions on different datasets with the same parameters.

The prediction is first performed on a random set of 11,000 letters. data sets of model loss were applied, along with accuracy and time as references. Due to the fact that text prediction is a classification problem, the cross entropy of binary classification was used as the accuracy in this experiment. The learning ability of the model was analyzed by model loss with accuracy as a reference.

In the model training, the time was recorded once at the beginning and end of each round, and the end time was subtracted from the start time to obtain the time of individual rounds. The computational efficiency of the model was determined based on the average length of training time after the results were available.

The next dataset chosen was Shakespeare's text, which is around 100,000 letters. When the model was officially used, the final results of model loss were used as a reference for accuracy and the total time was used for speed comparison.

\section{Results}

\subsection{Results of LSTM and E-LSTM with 11,000 random letters data set}

The results of LSTM and E-LSTM with 11,000 random letters data set are shown below:

\begin{figure}[ht]
	\centering
	\includegraphics[width=6in]{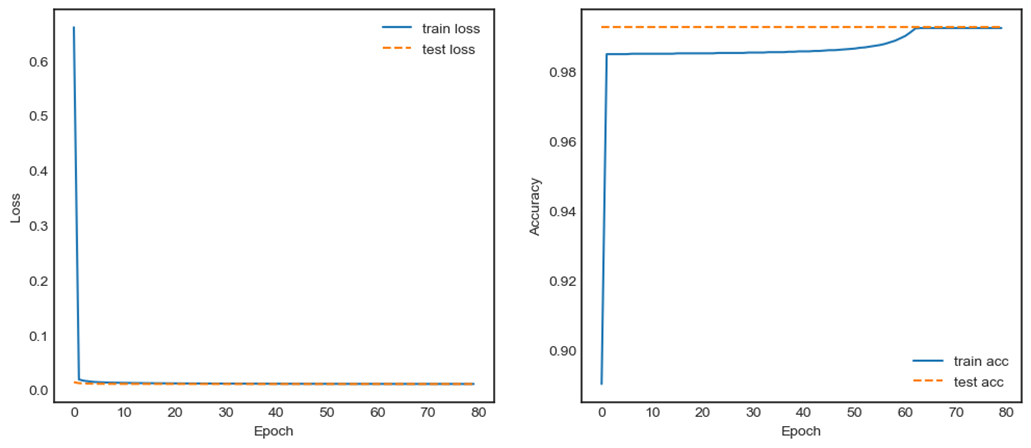}
	\caption{model loss and accuracy of LSTM with random 11,000 letters data set}
	\label{fig_LALSTM11000}
\end{figure}

Figure \ref{fig_LALSTM11000} shows the model loss and accuracy of LSTM with random 11,000 letters data set. There are 80 epochs for the training result. As can be seen from figure 4, the gradient of the model loss decrease is highest in the $2^{nd}$ epoch from 0.8399 to 0.0295, which means a decrease of 0.8104.

Then in the 3rd epoch, the model loss dropped to 0.02286, which is a decrease of 0.0066. As result, the gradient of the 3rd epoch decreases much less than the previous epoch. 

And then the gradient decrease more slowly. The gradient of model loss was 0.01193 at the $15^{th}$ epoch and 0.01099 at the $40^{th}$ epoch, which means the model loss only decreased by 0.00094 after 25 epochs.  Finally, the model loss decreased to 0.01073 at the final epoch.

From figure \ref{fig_LALSTM11000} above we can see that the accuracy of LSTM increase is highest in the $2^{nd}$ epoch from 0.8003 to 0.9789, which is an increase of 0.1786. After that, accuracy increased very less than before, which is from 0.9851 in the 3rd epoch to 0.9925 in the $55^{th}$ epoch. Then keep the accuracy at 0.9925 until the final epoch.

\begin{figure}[ht]
	\centering
	\includegraphics[width=6in]{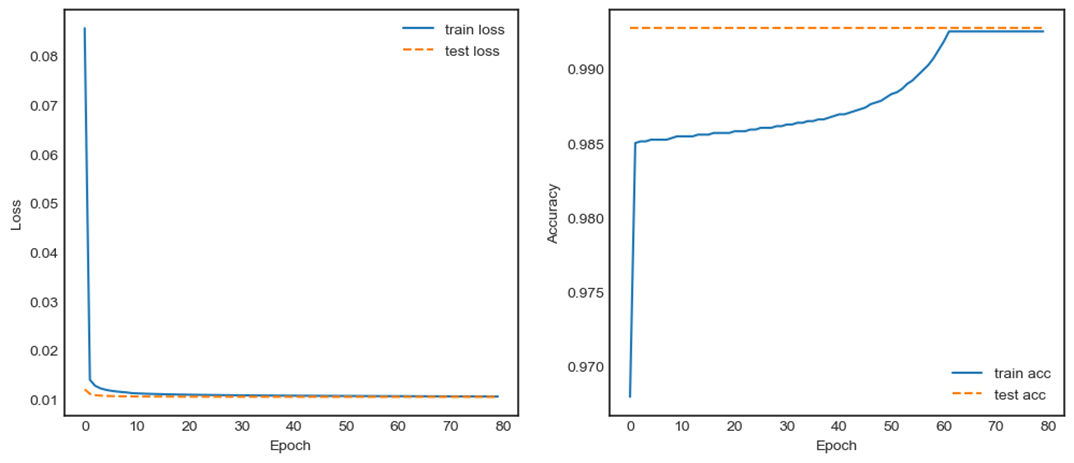}
	\caption{model loss and accuracy of E-LSTM with random 11,000 letters data set}
	\label{fig_LAELSTM11000}
\end{figure}

Figure \ref{fig_LAELSTM11000} shows the model loss and accuracy of E-LSTM with random 11,000 letters data set. There are 80 epochs for the training result. As can be seen from the figure, the gradient of the model loss decrease is highest in the $2^{nd}$ epoch from 0.08562 to 0.01391, which means a decrease of 0.07171.

Then in the $3^{rd}$ epoch, the model loss dropped to 0.01269, which is a decrease of 0.00122. The gradient of the third epoch decreases much less than the previous epoch. 

And then the gradient of model loss was 0.01097 at the $16^{th}$ epoch and 0.0105 at the final epoch, which means the model loss only decreased by 0.00092 in the other 64 epochs. In the other words, the model loss in the $16^{th}$ epoch is close to the final.

From figure \ref{fig_LAELSTM11000} above we can see that the accuracy of E-LSTM increase is highest in the $2^{nd}$ epoch from 0.9679 to 0.985, which is an increase of 0.0171. After that, the accuracy from 0.9851 in the $3^{rd}$ epoch to 0.9925 in the $62^{nd}$ epoch. Then keep the accuracy at 0.9925 until the final epoch.

\begin{figure}[ht]
	\centering
	\includegraphics[width=6in]{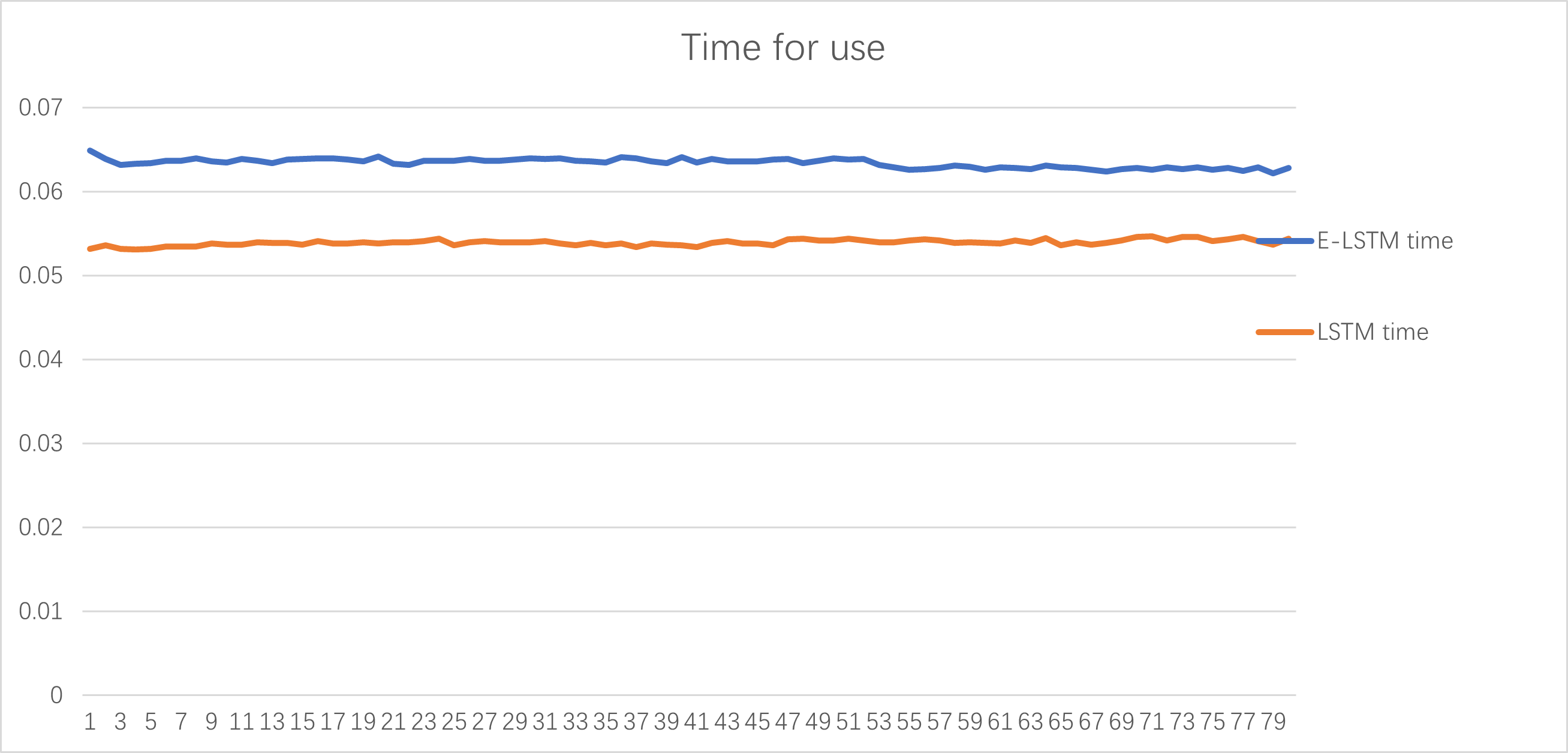}
	\caption{The time for use of E-LSTM and LSTM with random 11,000 letters data set}
	\label{fig_T11000}
\end{figure}

Figure \ref{fig_T11000} provides the time for use of E-LSTM and LSTM. The training time for the LSTM is between 0.05-0.06 minutes, which the average time being 0.05394 minutes. And the training time for E-LSTM is between 0.06-0.07 minutes, which the average time being 0.0634 minutes. Which means the E-LSTM needed 0.00946 minutes for training in every epoch, in the other words, the E-LSTM needed 17.53\% more time of LSTM in every epoch for training. In terms of overall efficiency, the training results of E-LSTM in the 33rd epoch have reached the training results of LSTM 80 epochs. It can be observed from figure 6 that the usage time of the two models does not fluctuate much in single epoch, therefore the overall time is about 2.1219 minutes for E-LSTM in 33rd epoch, and 4.3152 minutes for LSTM in 80th epoch times to achieve the same accuracy as E-LSTM. In other words, the overall usage time of LSTM is double that of E-LSTM. It is worth mentioning that this difference in accuracy is not fixed at the beginning. Experiments show that this value will gradually increase with the increase of training times.

In summary, these results show that this experiment verifies the theoretical part, which is the training time of E-LSTM is more than LSTM in an epoch, but the accuracy of E-LSTM is higher than LSTM. It can be seen from Figures \ref{fig_LALSTM11000} and \ref{fig_LAELSTM11000}. The $2^{nd}$ epoch model loss of E-LSTM is 0.01391 close to the $7^{th}$ epoch (0.01379) of LSTM training. When E-LSTM is training to the $33^{rd}$ epoch, the LSTM need to be training to the 80 epochs. In the other words, E-LSTM can achieve the result of 80 epochs of LSTM training with only 33 epochs of training, which takes only 17.5\% more time in every epoch.

\subsection{Results of LSTM and E-LSTM with Shakespeare's text data set}
The results of LSTM and E-LSTM with Shakespeare’s text data set are shown below:

\begin{figure}[ht]
	\centering
	\includegraphics[width=6in]{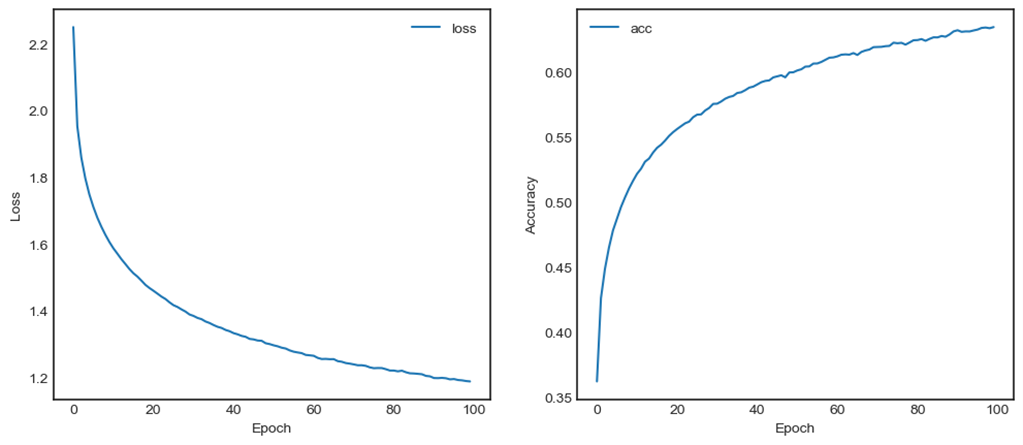}
	\caption{model loss and accuracy of LSTM with Shakespeare’s text data set}
	\label{fig_LALSTMskp}
\end{figure}

Figure \ref{fig_LALSTMskp} shows the model loss and accuracy of LSTM with Shakespeare’s text data set. There are 100 epochs for the training result. As can be seen from the figure, the gradient of the model loss decrease is highest in the $2^{nd}$ epoch from 2.2496 to 1.95287, which means a decrease of 0.29673.

Then in the $3^{rd}$ epoch, the model loss dropped to 1.85886, which is a decrease of 0.09401. As result, the gradient of the $3^{rd}$ epoch decreases much less than the previous epoch. 

And then the gradient decrease more slowly. The gradient of model loss was 1.60622 at the $10^{th}$ epoch and 1.46838 at the $20^{th}$ epoch, which means the model loss only decreased by 0.13784 in 10 epochs. Finally, the model loss decreased to 1.1886 in the final epoch.

From figure \ref{fig_LALSTMskp} above we can see that the accuracy of LSTM increase is highest in the $2^{nd}$ epoch from 0.362 to 0.4259, which is an increase of 0.0639. After that, accuracy increased more slowly than before, such as accuracy from 0.46 to 0.53 increase of 0.07 needed 9 epochs, which is from the $7^{th}$ epoch to the $13^{th}$ epoch. Then from 0.53 to 0.6 need 38 epochs, which from the $13^{th}$ to the $51^{st}$ epoch. Then the accuracy slowly increases to the $100^{th}$ epoch, which is 0.6341.

\begin{figure}[ht]
	\centering
	\includegraphics[width=6in]{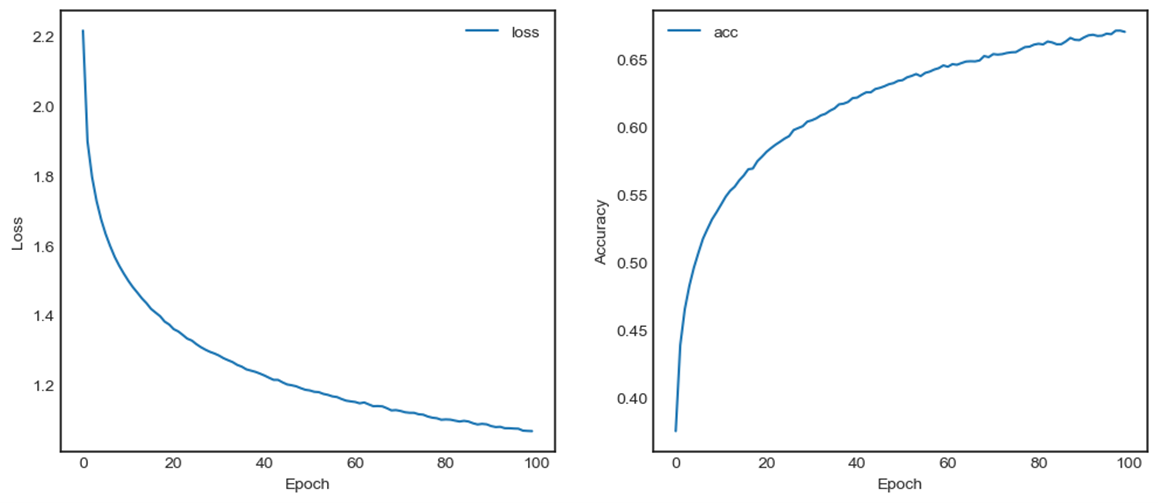}
	\caption{model loss and accuracy of ELSTM with Shakespeare’s text data set}
	\label{fig_LAELSTMskp}
\end{figure}

Figure \ref{fig_LAELSTMskp} shows the model loss and accuracy of E-LSTM with Shakespeare’s text data set. There are 100 epochs for the training result. As can be seen from the figure, the gradient of the model loss decrease is highest in the $2^{nd}$ epoch from 2.21438 to 1.89725, which means a decrease of 0.31713.

Then in the $3^{rd}$ epoch, the model loss dropped to 1.79548, which is a decrease of 0.10177. As result, the gradient of the $3^{rd}$ epoch decreases less than the $2^{nd}$ epoch. 

And then the gradient decreases more slowly as LSTM. The gradient of model loss was 1.59736 at the $7^{th}$ epoch and 1.46405 at the $13^{th}$ epoch, which means the model loss only decreased by 0.13331 in 5 epochs. After that, the model loss decreased to 1.18604 in the $50^{th}$ epoch. In the end, the model loss decreased to 1.06755 in the final epoch. 

From figure \ref{fig_LAELSTMskp} above we can see that the accuracy of E-LSTM increase is highest in the $2^{nd}$ epoch from 0.375 to 0.4384, which is an increase of 0.0634. After that, accuracy increased more slowly than before, such as accuracy from 0.4384 to 0.5365 increase of 0.0981 needed 7 epochs, which is from the $2^{nd}$ epoch to the $9^{th}$ epoch. Then from 0.5365 to 0.634 needs 41 epochs, which is from the $9^{th}$ to the $51^{st}$ epoch. Then the accuracy slowly increases to the $100^{th}$ epoch, which is 0.6697.

\begin{figure}[ht]
	\centering
	\includegraphics[width=6in]{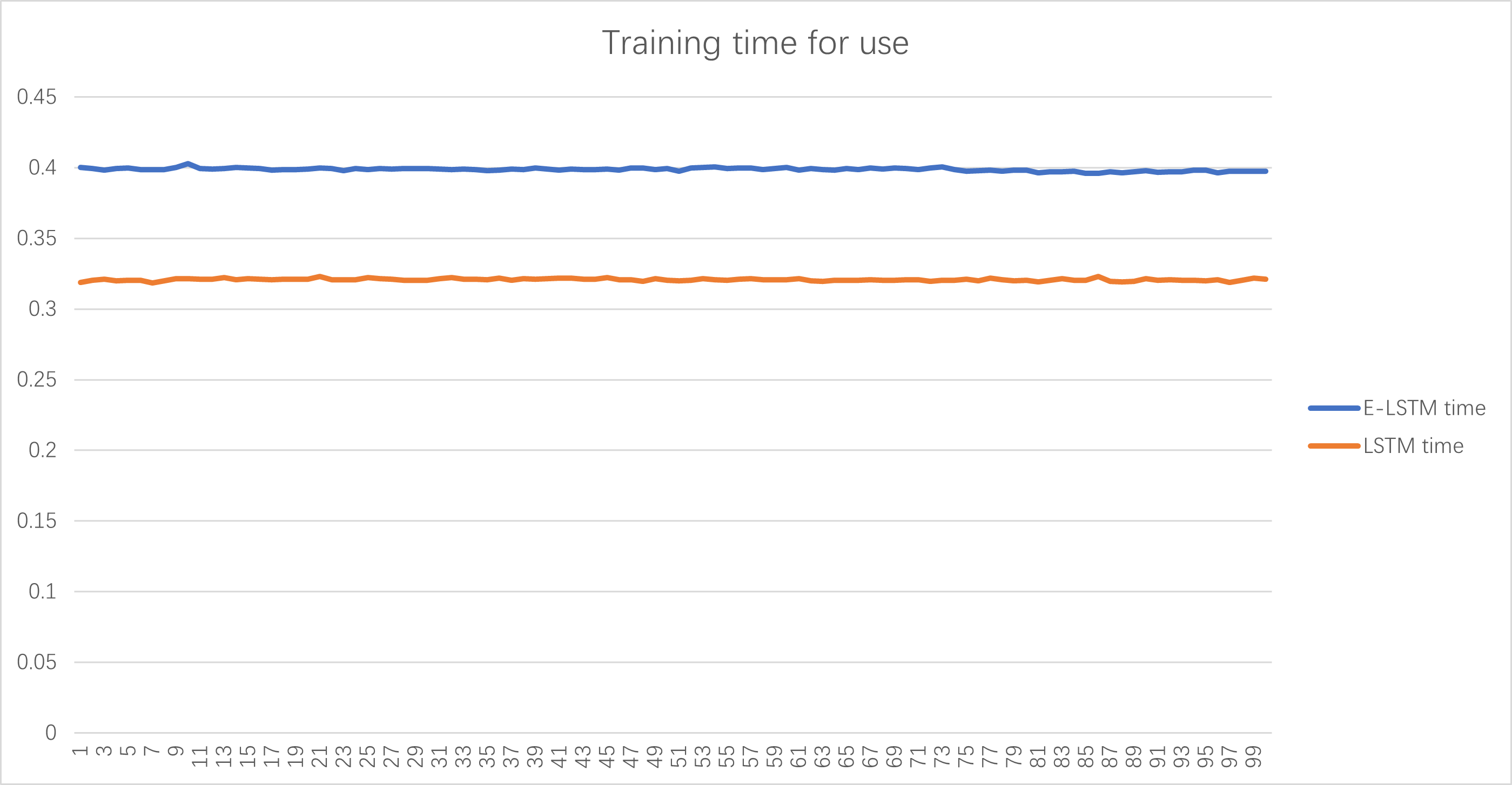}
	\caption{The training time for use of E-LSTM and LSTM with Shakespeare’s text data set}
	\label{fig_Tskp}
\end{figure}

Figure \ref{fig_Tskp} provides the training time for use of E-LSTM and LSTM. The training time for the LSTM is between 0.3-0.35 minutes, the average is 0.32074 minutes. And the training time for E-LSTM is around 0.4 minutes, the average is 0.399 minutes. Which means the E-LSTM needed 0.0778 more minutes for training in every epoch, in the other words, the E-LSTM needed 24.26\% more time of LSTM in every epoch for training.

In summary, these results show that this experiment is consistent with the theoretical part, which is the training time of E-LSTM is more than LSTM in an epoch, but the accuracy of E-LSTM is higher than LSTM. It can be seen from Figures \ref{fig_LAELSTMskp} and \ref{fig_Tskp}, the result in this experiment is a little different from the previous experiment. The value of the first 8 epochs of the LSTM and E-LSTM is close. The difference between the two models from the $9^{th}$ epoch. The model loss is 1.54103 in the $9^{th}$ epoch of E-LSTM, however, LSTM training to approximate the result requires 14 epochs. As the number of training increases, the gap between the LSTM and E-LSTM getting wider. The results have shown in figure \ref{fig_LALSTMskp} and \ref{fig_LAELSTMskp}, the E-LSTM obtains the model loss at $50^{th}$ epoch is equal to $100^{th}$ epoch of LSTM. This result also illustrates that E-LSTM takes only 24.26\% more time to reduce the training epoch by half.

In terms of overall efficiency, the training results of E-LSTM in the $50^{th}$ epoch have reached the training results of LSTM $100^{th}$ epochs. It can be observed from figure 9 that the usage time of the two models does not fluctuate much in single epoch, therefore the overall time is about 19.95 minutes for E-LSTM in $50^{th}$ epoch, and 32.07 minutes for LSTM in $100^{th}$ epoch times to achieve the same accuracy as E-LSTM. In other words, the overall usage time of LSTM is 1.607 times that of E-LSTM. It is worth mentioning that this difference in accuracy is not fixed at the beginning. Experiments show that this value will gradually increase with the increase of training times.

\section{Conclusions}
In conclusion, this research proposes a new LSTM called E-LSTM. Combining the advantages of LSTM and ELM by adding an ELM 'gate' to the LSTM, the ELM calculates the results with certain accuracy and participates in the LSTM for iteration, which can reduce the number of LSTM iterations. Experimentally, the results of 11,000 random letter data set and Shakespeare’s text data set fitting the theoretical. It can be observed from the result E-LSTM has higher training accuracy in single epoch. In general, the overall usage time of LSTM is more than 1.6 times that of E-LSTM, and this value will increase by increase of training epochs.


\clearpage
\appendix
\section{Appendix}
\begin{figure}[ht]
	\centering
	\includegraphics[width=4.8in]{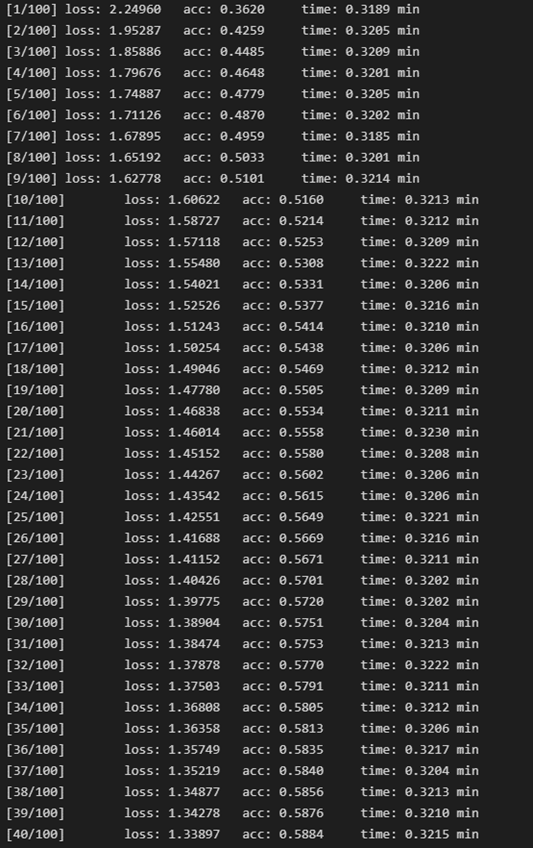}
	\caption{Results of LSTM with Shakespeare's text data set\_1}
	\label{fig_iLSTMskp1}
\end{figure}

\begin{figure}[ht]
	\centering
	\includegraphics[width=5in]{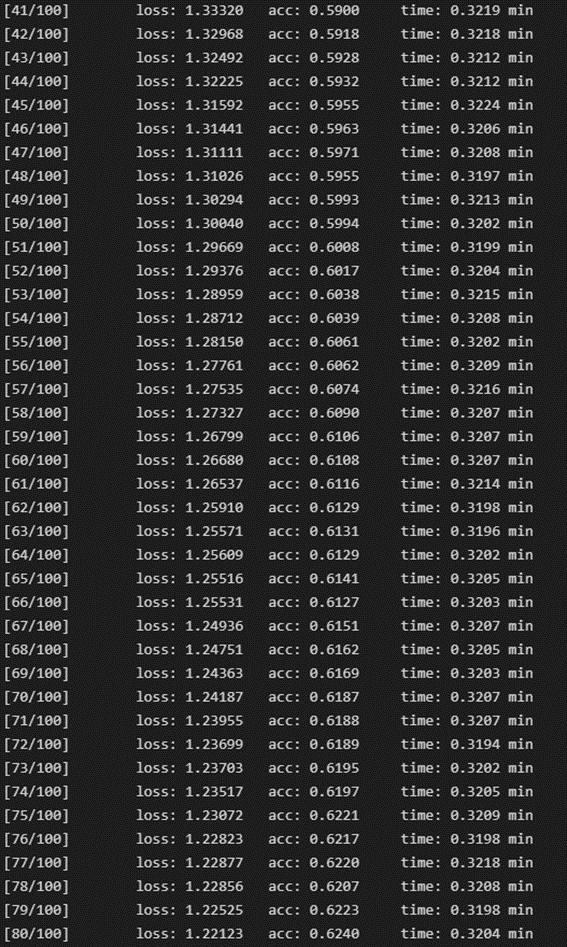}
	\caption{Results of LSTM with Shakespeare's text data set\_2}
	\label{fig_iLSTMskp2}
\end{figure}

\begin{figure}[ht]
	\centering
	\includegraphics[width=5in]{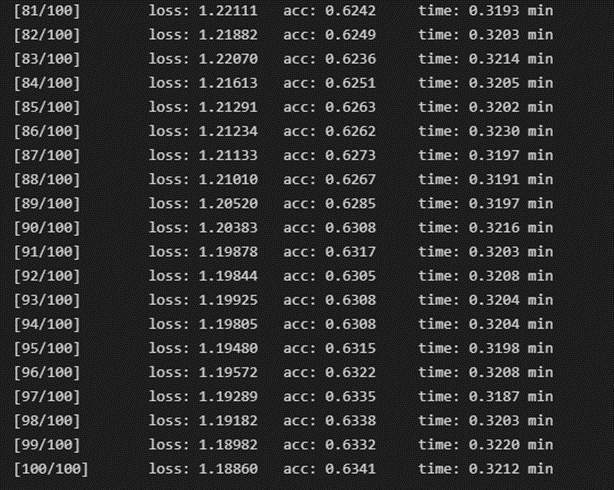}
	\caption{Results of LSTM with Shakespeare's text data set\_3}
	\label{fig_iLSTMskp3}
\end{figure}

\begin{figure}[ht]
	\centering
	\includegraphics[width=5in]{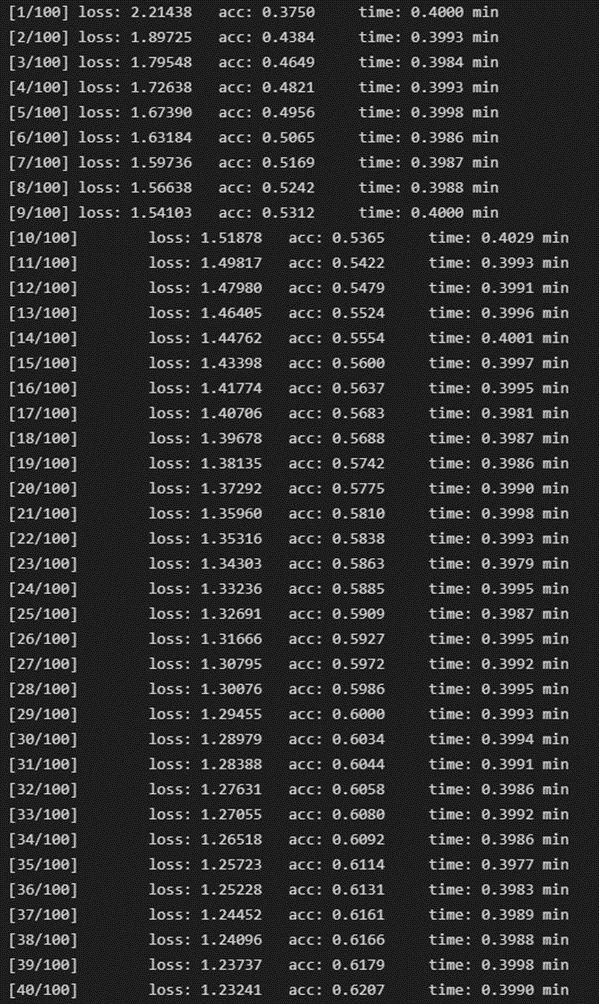}
	\caption{Results of E-LSTM with Shakespeare's text data set\_1}
	\label{fig_iELSTMskp1}
\end{figure}

\begin{figure}[ht]
	\centering
	\includegraphics[width=5in]{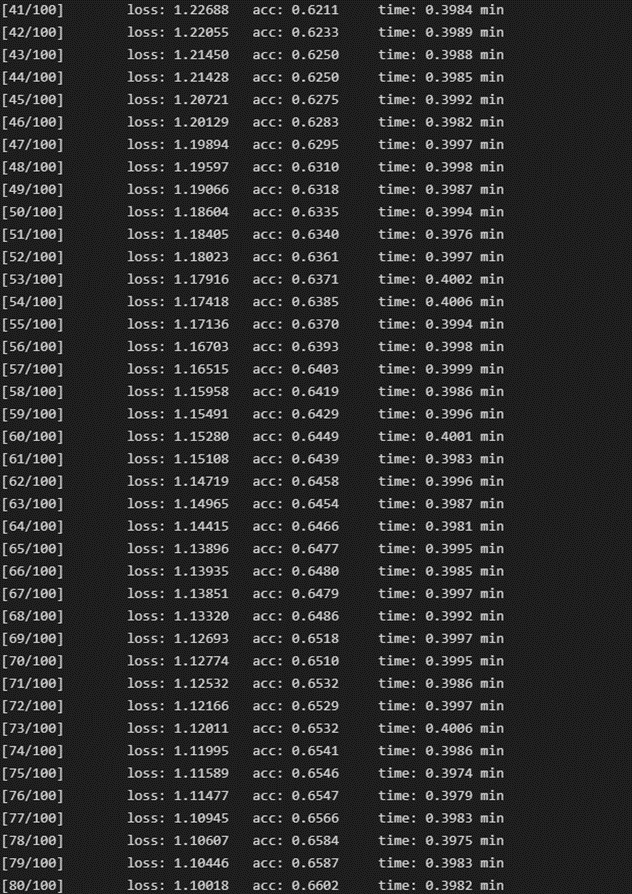}
	\caption{Results of E-LSTM with Shakespeare's text data set\_2}
	\label{fig_iELSTMskp2}
\end{figure}

\begin{figure}[ht]
	\centering
	\includegraphics[width=5in]{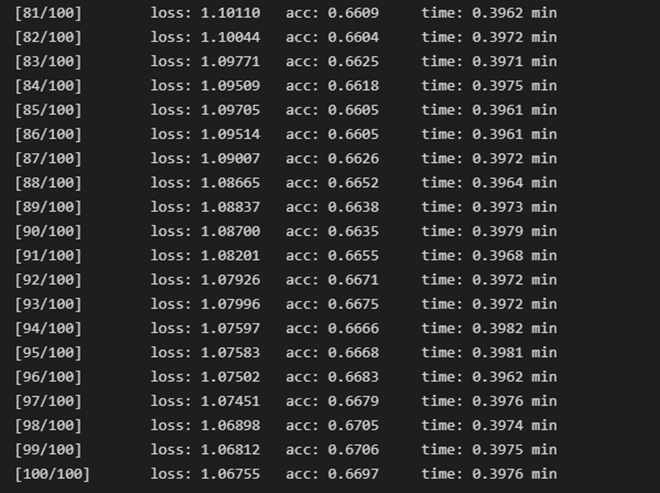}
	\caption{Results of E-LSTM with Shakespeare's text data set\_3}
	\label{fig_iELSTMskp3}
\end{figure}

\end{document}